\documentclass[sigconf,review=false,anonymous=false]{acmart}
\usepackage{multirow,subfigure,pifont,hyperref,amsfonts}
\newcommand{\tabincell}[2]{\begin{tabular}{@{}#1@{}}#2\end{tabular}}
\newcommand{\cmark}{\ding{51}}%
\newcommand{\xmark}{\ding{55}}%
\acmConference{ACM Multimedia}{October 12--16, 2020}{Seattle}
\AtBeginDocument{%
  \providecommand\BibTeX{{%
    \normalfont B\kern-0.5em{\scshape i\kern-0.25em b}\kern-0.8em\TeX}}}

\copyrightyear{2020}
\acmYear{2020}
\setcopyright{acmlicensed}\acmConference[MM '20]{Proceedings of the 28th
ACM International Conference on Multimedia}{October 12--16, 2020}{Seattle,
WA, USA}
\acmBooktitle{Proceedings of the 28th ACM International Conference on
Multimedia (MM '20), October 12--16, 2020, Seattle, WA, USA}
\acmPrice{15.00}
\acmDOI{10.1145/3394171.3413922}
\acmISBN{978-1-4503-7988-5/20/10}

\begin{document}
\fancyhead{}
\title{LSOTB-TIR: A Large-Scale High-Diversity Thermal Infrared Object Tracking Benchmark}


\author{Qiao Liu}
\author{Xin Li}
\affiliation{%
  \institution{Harbin Institute of Technology, Shenzhen}
}
\email{liuqiao.hit@gmail.com} 

\author{Zhenyu He}
\authornote{Corresponding author.}
\affiliation{%
  \institution{Harbin Institute of Technology, Shenzhen}
}
\affiliation{%
  \institution{Peng Cheng Laboratory}
}
\email{zhenyuhe@hit.edu.cn}
\author{Chenglong Li}
\affiliation{%
  \institution{Anhui University}
}
\email{chenglongli@ahu.edu.cn}

\author{Jun Li}
\author{Zikun Zhou}
\author{Di Yuan}
\affiliation{%
  \institution{Harbin Institute of Technology, Shenzhen}
}

\author{Jing Li}
\author{Kai Yang}
\author{Nana Fan}
\affiliation{%
  \institution{Harbin Institute of Technology, Shenzhen}
}

\author{Feng Zheng}
\affiliation{%
  \institution{Southern University of Science and Technology}
}
\email{zfeng02@gmail.com}

%
%
%
%
%
%


\begin{abstract}
 In this paper, we present a Large-Scale and high-diversity general Thermal InfraRed (TIR) Object Tracking Benchmark, called LSOTB-TIR, which consists of an evaluation dataset and a training dataset with a total of 1,400 TIR sequences and more than 600K frames.
 We annotate the bounding box of objects in every frame of all sequences and generate over 730K bounding boxes in total.
 To the best of our knowledge, LSOTB-TIR is the largest and most diverse TIR object tracking benchmark to date.
 To evaluate a tracker on different attributes, we define 4 scenario attributes and 12 challenge attributes in the evaluation dataset.
 By releasing LSOTB-TIR, we encourage the community to develop deep learning based TIR trackers and evaluate them fairly and comprehensively.
 We evaluate and analyze more than 30 trackers on LSOTB-TIR to provide a series of baselines, and the results show that deep trackers achieve promising performance.
 Furthermore, we re-train several representative deep trackers on LSOTB-TIR, and their results demonstrate that the proposed training dataset significantly improves the performance of deep TIR trackers. Codes and dataset are available at \url{https://github.com/QiaoLiuHit/LSOTB-TIR}.
\end{abstract}

\ccsdesc[500]{Computing methodologies~Tracking}
\ccsdesc[500]{Computing methodologies~Image representations}

\keywords{thermal infrared dataset, thermal infrared object tracking, deep representation learning}


\maketitle

\begin{figure}[t]
	\begin{center}
		\includegraphics[width=0.47\textwidth]{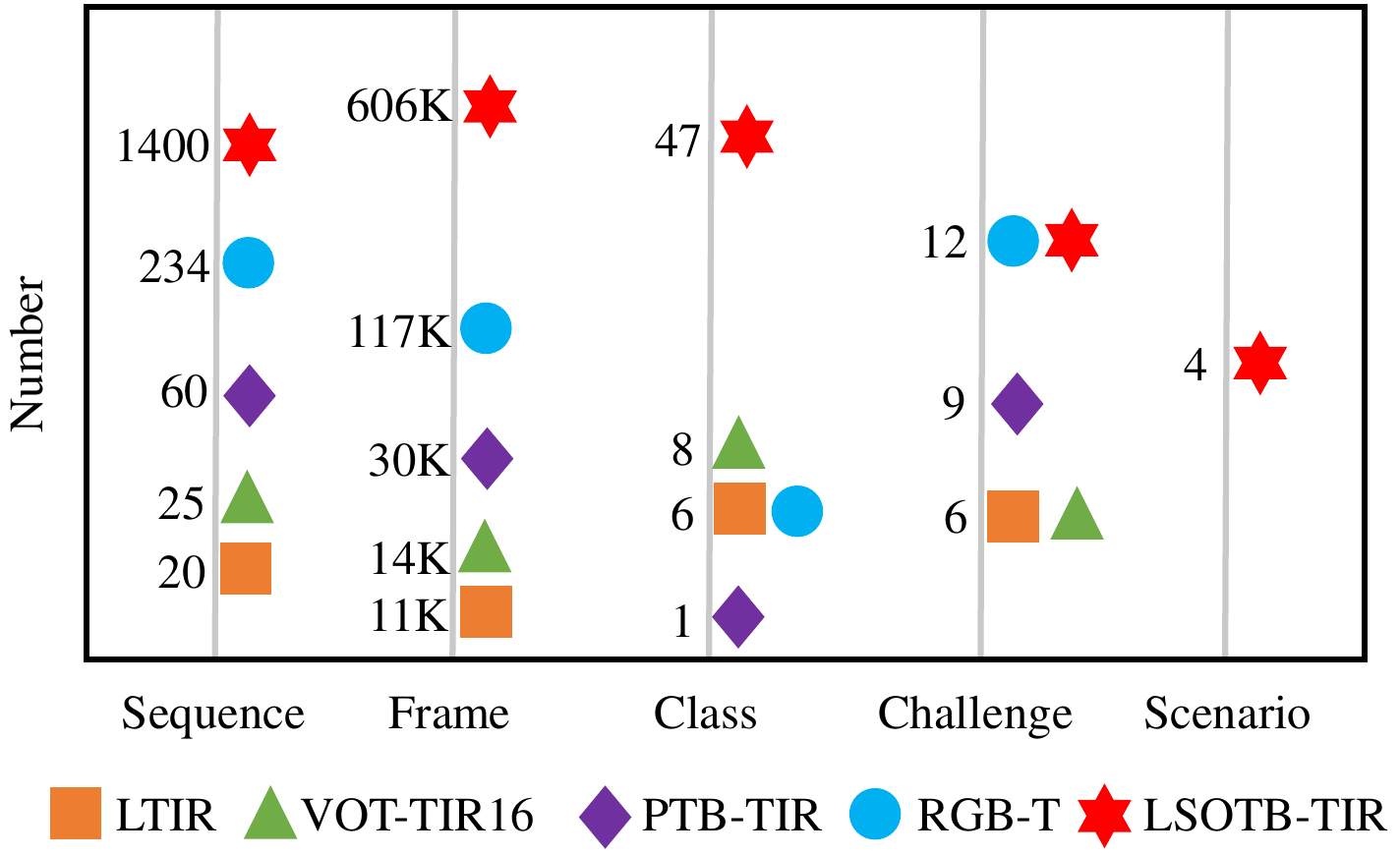}
	\end{center}
	\caption{Comparison of currently widely used TIR object tracking benchmarks, including LTIR~\cite{LTIR}, VOT-TIR16~\cite{VOT-TIR2016}, PTB-TIR~\cite{PTB-TIR}, RGB-T~\cite{RGB-T}, and LSOTB-TIR. Among these benchmarks, LSOTB-TIR is the largest and most diverse. Noting that only LSOTB-TIR contains the scenario attribute. }
	\label{comparison-benchmark}
\end{figure}

\section{Introduction}
TIR object tracking is an important task in artificial intelligence.
Given an initial position of a TIR object in the first frame, TIR object tracking is to locate the object in the rest of the frames of a sequence.
With the popularization of civilian thermal imaging devices, TIR object tracking receives more and more attention as a crucial intelligent vision technology.
It is widely used in video surveillance, maritime rescue, and driver assistance at night~\cite{TC} since it can track the object in total darkness.
In the past several years, some TIR object tracking methods~\cite{HSSNet,DSST-tir,MCFTS,DSLT,LMSCO,ECO-tir} are proposed.
Despite much progress, TIR object tracking faces many unsolved problems, such as distractor, intensity variation, and thermal crossover~\cite{PTB-TIR}.

Evaluating a tracker fairly and comprehensively on a benchmark is crucial to the development of TIR object tracking.
However, currently widely used TIR object tracking benchmarks, e.g., LTIR~\cite{LTIR}, VOT-TIR16~\cite{VOT-TIR2016}, and PTB-TIR~\cite{PTB-TIR} suffer from the following drawbacks that make them less effective in conducting a fair and comprehensive evaluation.
First, their scale is too small to make an effective evaluation because a tracker can easily overfit to a small dataset using the parameter fine-tuning.
Second, they have too few kinds of objects, \textit{e.g.}, PTB-TIR only contains pedestrian objects, which cannot provide an evaluation on general TIR objects.
Third, they only have a few tracking scenarios and challenges, which does not meet the requirements of real-world applications.
Therefore, it is imperative to build a larger and more diverse TIR object tracking benchmark.

\begin{figure}[t]
	\begin{center}
		\subfigure[Distribution of intra-class TIR objects]{\includegraphics[width=0.23\textwidth]{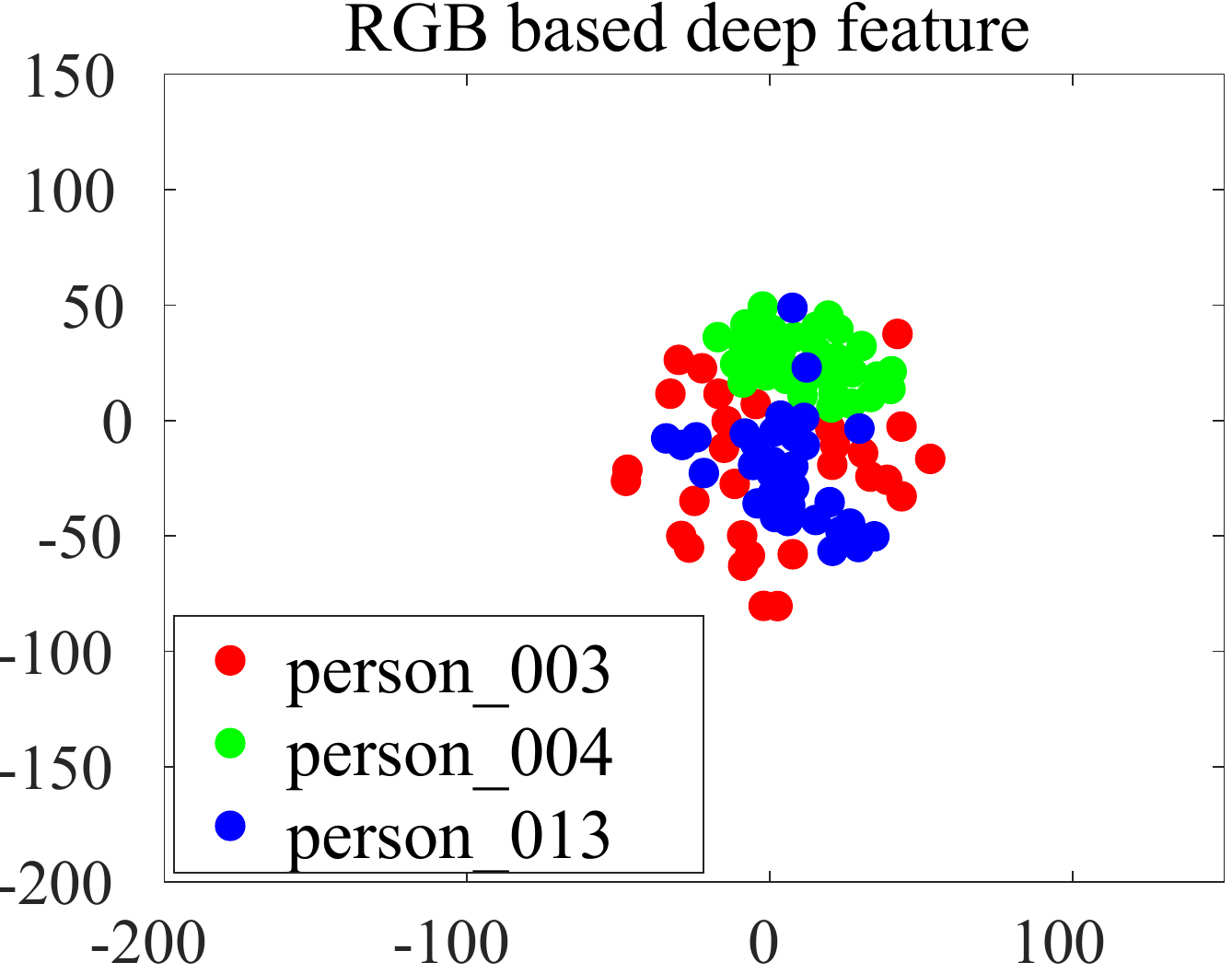}
			\includegraphics[width=0.23\textwidth]{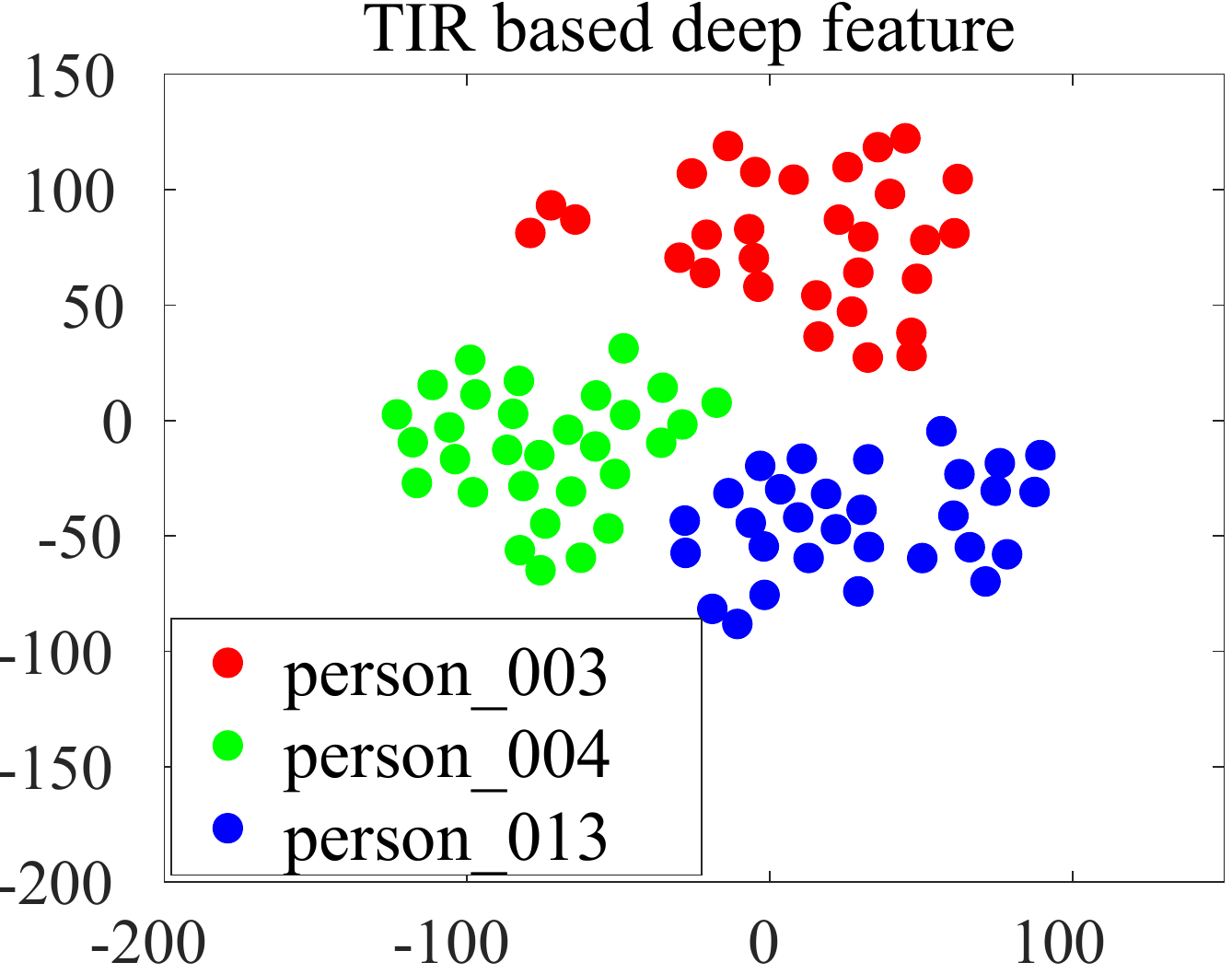}}
		\subfigure[Distribution of inter-class TIR objects]{\includegraphics[width=0.23\textwidth]{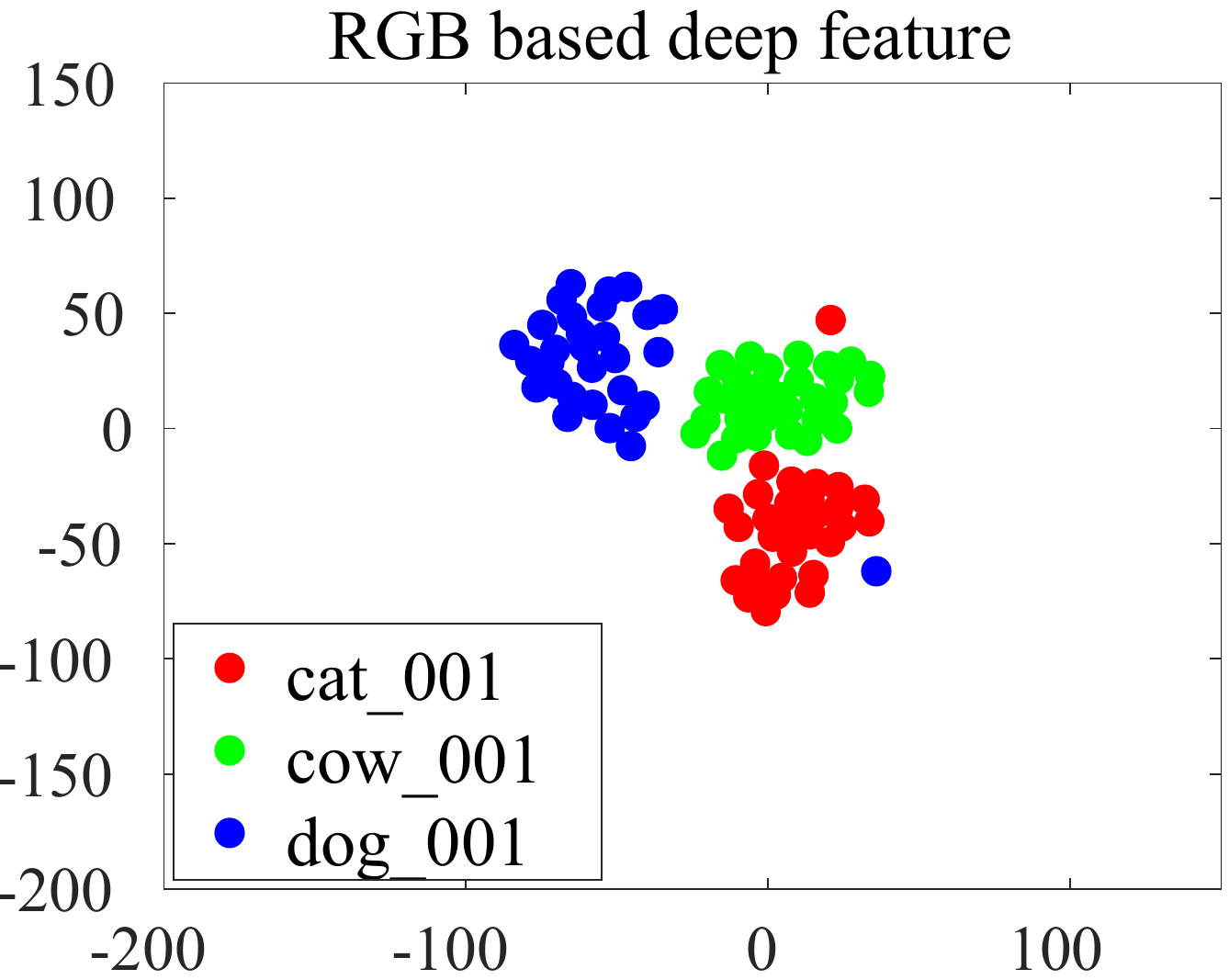}
			\includegraphics[width=0.23\textwidth]{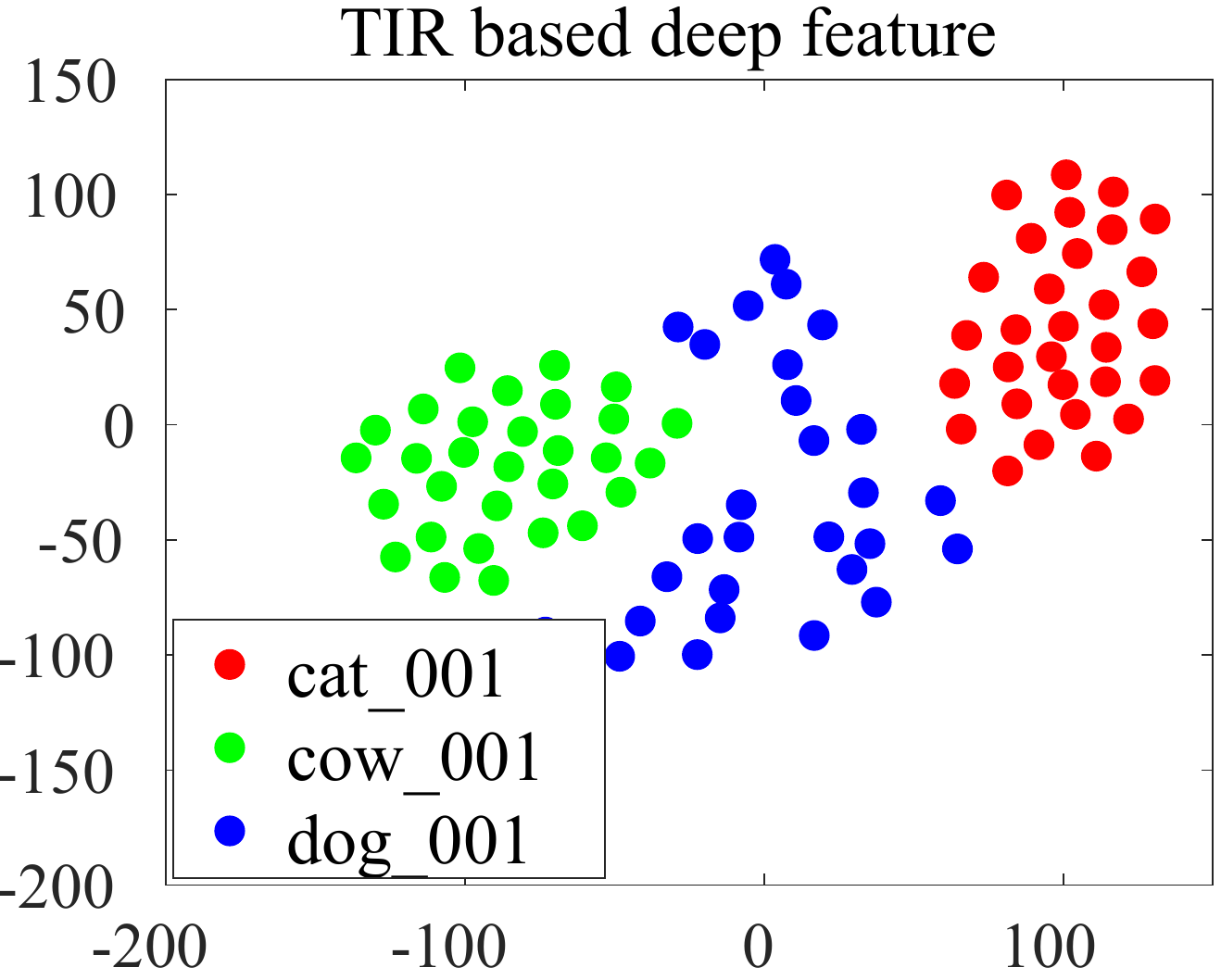}}
	\end{center}
	\caption{Comparison of the RGB based deep feature and the TIR based deep feature using the t-SNE visualized method~\cite{t-SNE}. The RGB and TIR based deep features are extracted from the backbone network of two CFNets~\cite{CFNet}, which are trained on an RGB dataset (VID~\cite{ILSVRC}) and LSOTB-TIR, respectively. We randomly choose 30 objects from each sequence in LSOTB-TIR. (a) All objects belong to the person class but come from different sequences. We can see that the TIR based deep feature can recognize the differences between intra-class TIR objects, which is curial to distinguish distractors in TIR tracking. (b) All objects belong to different classes. The TIR based deep feature can separate inter-class objects more effectively than the RGB based deep feature. Noting that each point in the figure denotes an object.}
	\label{feature-comparison}
\end{figure}

Recently, motivated by the success of deep learning in most visual tasks, several attempts~\cite{MCFTS,HSSNet,LMSCO,MLSSNet} incorporate deep feature models for TIR object tracking and achieve some success.
However, the used deep feature models are learned from RGB images, and we find by experiments that these RGB based deep feature models are less effective in representing TIR objects, as shown in Fig.~\ref{feature-comparison}.
Different from RGB images, TIR images do not have color information and lack texture features, hence it is crucial to use fine-grained features, such as local contour and structure, to distinguish objects in TIR images.
Furthermore, we find by experiments that learning TIR-specific deep features for representing further promotes the performance of TIR object tracking.
However, the lacking of a large-scale TIR dataset for deep model training hinders the advantage of the deep leaning in TIR object tracking.

To address the above-mentioned issues, we develop a large-scale and high-diversity TIR object tracking benchmark, called LSOTB-TIR,
which consists of an evaluation dataset and a training dataset with a total of 1,400 TIR sequences and more than 600K frames.
We annotate the bounding box of objects in every frame of all sequences and generate more than 730K bounding boxes in total.
We carefully select 120 sequences with 22 object classes and more than 82K frames as the evaluation dataset, which is larger and more diverse than existing TIR datasets.
To understand the strengths and weaknesses of a tracker on specific attributes, we define 4 scenario attributes and 12 challenge attributes for attribute-based evaluation.
The training dataset contains 1,280 sequences with 47 object classes and over 650K bounding boxes, which is used for learning TIR-specific deep features.
In addition to the benchmark, we evaluate more than 30 trackers on LSOTB-TIR, and provide a detailed analysis.
The results show that deep trackers achieve promising performance.
Moreover, we re-train several representative deep trackers using the proposed training dataset, and their results on three benchmarks demonstrate that the proposed TIR training dataset significantly improves the performance of deep TIR trackers.

The contributions of this paper are three-fold:
\begin{itemize}
	\item We propose a large-scale TIR object tracking benchmark, LSOTB-TIR, with high-quality annotations using a self-designed semi-automatic label tool.
	LSOTB-TIR is currently the largest and most diverse TIR object tracking benchmark, which consists of the training and evaluation datasets with the richest object classes, scenarios, and challenges.
	\item We conduct extensive evaluation experiments with more than 30 trackers on LSOTB-TIR and provide a series of comparative analyses.
	The results show that deep trackers achieve promising results and have the potential to obtain better performance.
	\item We re-train several representative deep trackers on LSOTB-TIR, and their results on three benchmarks demonstrate that the proposed training dataset significantly improves the performance of deep TIR trackers.
\end{itemize}

\begin{table*}[t]
	\centering
	\caption{Comparison of the proposed LSOTB-TIR benchmark with other TIR object tracking benchmarks.}
	\small
	\begin{tabular}{l|rrrrrrrrrr}
		\toprule
		Benchmarks &\tabincell{c}{Num. of \\ sequences} & \tabincell{c}{Max \\ frames} & \tabincell{c}{Min\\ frames} &\tabincell{c}{Mean\\ frames}  &\tabincell{c}{Total \\frames} & \tabincell{c}{Frame \\ rates} &\tabincell{c}{Object \\ classes} & \tabincell{c}{Num. of \\challenges}  &\tabincell{c}{Scenario\\ attributes} & \tabincell{c}{Training\\ dataset} \\
		\midrule
		OSU~\cite{OSU}  &6  &2,031  &601  &1,424  &8K  & 30 fps  &1 &n/a &  \xmark   & \xmark \\
		PDT-ATV~\cite{PDT-ATV} &8  &775  &77  &486  &4K  & 20 fps & 3 &n/a & \xmark & \xmark \\
		BU-TIV~\cite{TIV} &16 &26,760 &150 &3,750 &60K &30 fps &5 &n/a & \xmark & \xmark \\
		LTIR~\cite{LTIR}  &20 &1,451 & 71 &563 &11K & 30 fps &6 & 6 & \xmark & \xmark \\
		VOT-TIR16~\cite{VOT-TIR2016} &25 &1,451 & 71 &555 &14K &30 fps &8 &6 & \xmark & \xmark \\
		PTB-TIR~\cite{PTB-TIR} &60 &1,451 &50 &502 & 30K & 30 fps & 1 &9 & \xmark & \xmark \\
		RGB-T~\cite{RGB-T} &234 &4,000 &45 &500 &117K &30 fps &6 &12 & \xmark & \xmark \\
		\midrule
		LSOTB-TIR (train.) &1,280  &3,056 & 47 & 410 &524K  & 30 fps &47 & 1  & \multirow{2}{*}{\cmark } & \multirow{2}{*}{\cmark} \\
		LSOTB-TIR (eval.) &120  &2,110 & 105 & 684 &82K  & 30 fps &22 & 12  &  &\\
		\bottomrule
	\end{tabular}\label{benchmark}
\end{table*}

\section{Related Work}

\subsection{TIR trackers}
\label{tirtracker}
In the past decade, some TIR object tracking algorithms have been proposed to handle various challenges.
These algorithms can be roughly divided into two categories: conventional TIR trackers and deep TIR trackers.

\vspace{0mm}
\noindent{\textbf{Conventional TIR trackers.}} Conventional TIR object tracking methods usually combine a conventional machine learning method with a hand-crafted feature for handling various challenges.
To adapt the appearance variation of the object, Venkataraman \textit{et al}.~\cite{HBA} propose to online learn a robust intensity histogram based appearance model using adaptive Kalman filtering, while TBOOST~\cite{TBOOST} maintains a dynamic MOSSE filter~\cite{MOSSE} set using a continuously switching mechanism according to appearance variation.
Demir \textit{et al}.~\cite{PBCD} use a part-based matching method, which integrates the co-difference feature of multiple parts to overcome the partial deformation.
To obtain more effective representations, DSLT~\cite{DSLT} combines gradient histograms with motion features and then is used in a Structural Support Vector Machine (SSVM)~\cite{struck} for TIR object tracking.
Observing that TIR images do not have color information and lack sharp edges, Berg \textit{et al}.~\cite{ABCD} propose a distribution field representation~\cite{DFT} based matching algorithm for TIR object tracking.
Despite much progress, these trackers are limited by the hand-crafted feature representation.

\vspace{0mm}
\noindent{\textbf{Deep TIR trackers.}} Inspired by the success of deep learning in visual tracking, several works introduce the Convolution Neural Network (CNN) to improve the performance of TIR trackers.
These methods can be roughly divided into two categories, deep feature based TIR trackers and matching based deep TIR trackers.
Deep feature based TIR trackers often use a pre-trained CNN for feature extraction and then integrates the deep feature into conventional tracking frameworks.
For example, Gundogdu \textit{et al}.~\cite{DSST-tir} train a classification network on a small TIR dataset to extract the deep feature of the TIR object and then combine it with the DSST tracker~\cite{DSST} for TIR object tracking.
MCFTS~\cite{MCFTS} combines multiple convolutional features of VGGNet~\cite{VGGNet} with the Correlation Filters (CFs)~\cite{KCF} to construct an ensemble TIR tracker.
Gao \textit{et al}.~\cite{LMSCO} combine deep appearance features~\cite{VGGNet} and deep motion features~\cite{deepmotion} with SSVM for TIR object tracking.
ECO-stir~\cite{ECO-tir} trains a Siamese network on synthetic TIR images to extract TIR features and then integrates them into the ECO~\cite{ECO} tracker.
Li \textit{et al}.~\cite{MaskTIR} propose a mask sparse representation deep appearance model with the particle filter framework for TIR object tracking.
Matching based deep TIR trackers cast the tracking as a matching problem and usually off-line train a matching network for online tracking.
For example, Li \textit{et al}.~\cite{HSSNet} train a spatial variation aware matching network by introducing a spatial attention mechanism for TIR object tracking.
Liu \textit{et al}.~\cite{MLSSNet} propose a multi-level similarity based matching network using a semantic similarity module and a complementary structural similarity module for TIR object tracking.
However, most of these deep models are learned from RGB images, which do not learn specific patterns of TIR images and hence are less effective in representing TIR objects.

\subsection{TIR object tracking benchmarks}
\label{tirbenchmarks}
To evaluate TIR trackers, there are several widely used benchmarks~\cite{OSU,PDT-ATV,TIV,LTIR,VOT-TIR2015,VOT-TIR2016,PTB-TIR,RGB-T}.
In the following, we introduce these benchmarks briefly.

\vspace{0mm}
\noindent{\textbf{OSU.}} OSU~\cite{OSU} is a TIR and RGB image fusion dataset, which can be used for TIR object tracking.
This dataset contains only pedestrian objects and all 6 videos are captured by a low-resolution TIR camera in a static background.

\vspace{0mm}
\noindent{\textbf{PDT-ATV.}} PDT-ATV~\cite{PDT-ATV} is a simulative aerial TIR object tracking and detection dataset, which contains 8 sequences captured from a low frame rate and low-resolution TIR camera. The dataset does not have attribute labels and the tracking objects are dim and small.

\vspace{0mm}
\noindent{\textbf{BU-TIV.}} BU-TIV~\cite{TIV} is used for several TIR visual tasks, including object tracking, counting, and group motion estimation. This dataset contains 16 sequences and more than 60K frames with a high resolution.

\vspace{0mm}
\noindent{\textbf{LTIR.}} LTIR~\cite{LTIR} is the first standard TIR object tracking benchmark which contains 20 sequences with 6 object classes and an evaluation toolkit. This benchmark is adopted by a TIR object tracking competition, VOT-TIR15~\cite{VOT-TIR2015}.

\vspace{0mm}
\noindent{\textbf{VOT-TIR16.}} VOT-TIR16~\cite{VOT-TIR2016} is extended from VOT-TIR15. It contains 25 sequences with 8 object classes and is more challenging than VOT-TIR15. It has 6 challenge subsets that can be used to evaluate a tracker on specific attributes.

\vspace{0mm}
\noindent{\textbf{PTB-TIR.}} PTB-TIR~\cite{PTB-TIR} focuses on TIR pedestrian tracking, which contains 60 sequences and 9 attribute challenges. The dataset is collected from a variety of devices but lacks the division of different scenario attributes.

\vspace{0mm}
\noindent{\textbf{RGB-T.}} RGB-T~\cite{RGB-T} is a multi-modal tracking benchmark that contains RGB and TIR videos on the same scene simultaneously. The TIR videos are captured from a single low-resolution TIR camera and can also be used for TIR object tracking.

\begin{figure*}[htbp]
	\centering
    	\includegraphics[width=0.99\textwidth]{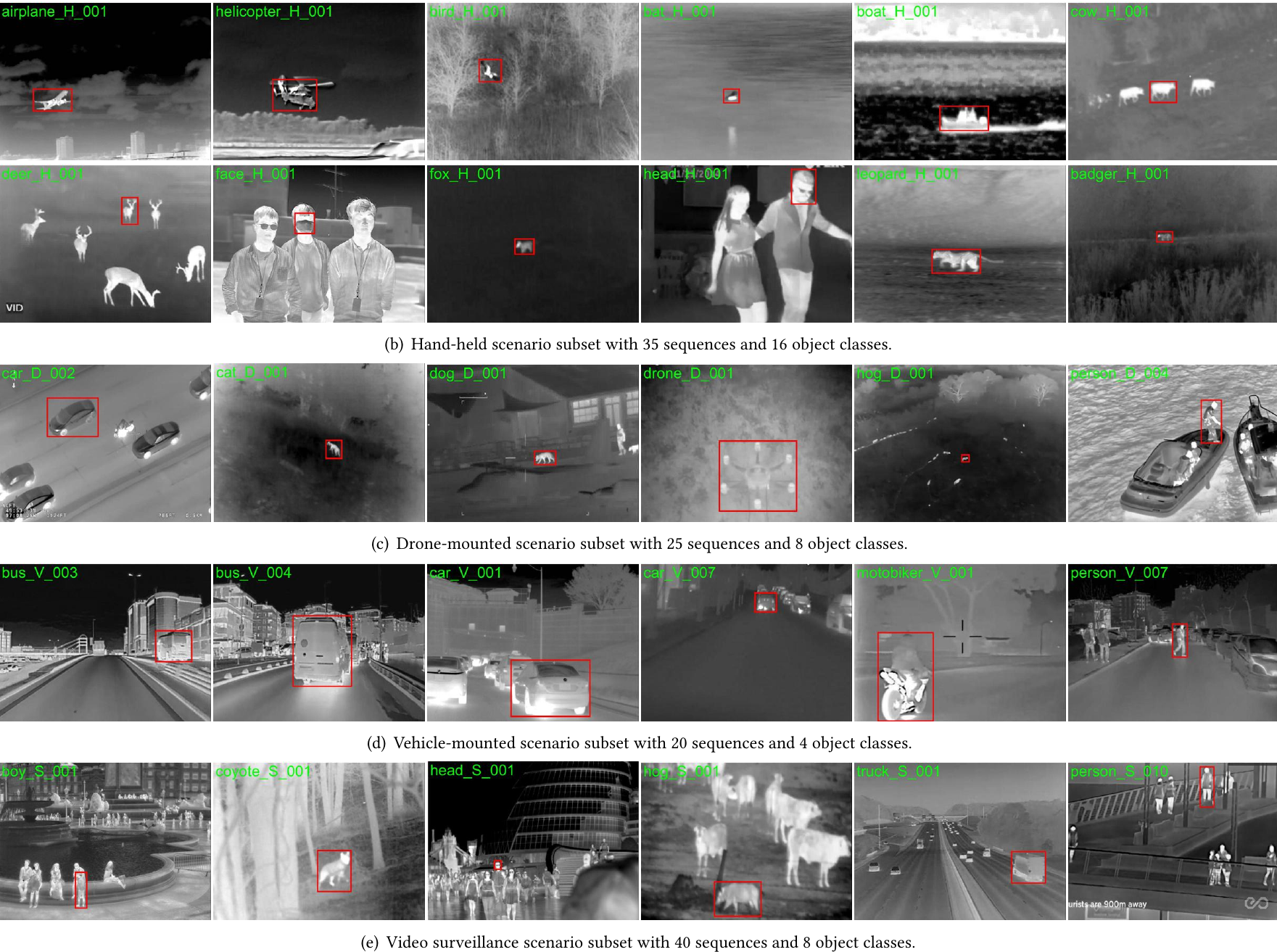}
	\caption{Examples of the proposed evaluation subset of LSOTB-TIR. It contains 4 non-overlapped scenario subsets with a total of 120 sequences and 22 object classes. }
	\label{evaluation-dataset}
\end{figure*}

\begin{table*}[t]
	\centering
	\caption{Definition of 4 scenarios and 12 challenges on LSOTB-TIR.}
	\small
	\begin{tabular}{ll|ll}
		\toprule
		Scenario & Definition & Scenario & Definition \\
		\midrule
		VS & The videos come from a surveillance camera. & DS & The videos are captured from a drone-mounted camera.\\
		HH & The videos are shotted from a hand-held camera. & VM & The videos come from a vehicle-mounted camera.\\
		\midrule
		\midrule
		Challenge & Definition & Challenge & Definition\\
		\midrule
		TC & Two same intensity targets cross each other. & IV & The target intensity is changed during tracking. \\
		DIS &  Existing the intra-class object near the target. & BC & The background has a similar appearance to the target.\\
		DEF & The target is deformable during tracking. & OCC & The target is partly or fully occluded during tracking. \\
		OV & The target partly or fully leaves the image. & SC & The ratio of the target size is out of the range [0.5, 2]. \\
		FM & The target moves more than 20 pixels.  &MB   & The target is blurry due to the target or camera motion. \\
		LR & The target size is lower than 800 pixels. & ARV & The target aspect ratio is out of the range [0.5, 2]. \\
		\bottomrule
	\end{tabular}\label{attribute-defination}
\end{table*}

Although these benchmarks are widely used, they suffer from several problems, such as the small-scale, limited object classes, scenarios and challenges, and lack of training dataset.
To solve these issues, we present a large-scale and high-diversity TIR object tracking benchmark, LSOTB-TIR, consisting of an evaluation dataset and a training dataset.
The evaluation dataset contains 120 sequences with more than 82K frames, 22 classes, 4 scenario subsets, and 12 challenge subsets, which is more diverse than these benchmarks.
The training dataset contains 1,280 sequences with more than 650K bounding boxes and 47 object classes.
Table~\ref{benchmark} compares the proposed benchmark with existing TIR object tracking benchmarks. More comparisons with existing RGB tracking benchmarks are shown in the \textbf{supplementary material}.

\section{Proposed LSOTB-TIR Benchmark}
In this section, we describe details of the proposed TIR object tracking benchmark, LSOTB-TIR. We first introduce  TIR videos collection and processing in Section~\ref{datacollection} and then we show how to annotate the sequence in Section~\ref{annoation}. Finally, we define attributes of a sequence in Section~\ref{attributes}.

\subsection{Data collection and processing}
\label{datacollection}
Our goal is to provide a large-scale and high-diversity general TIR object tracking benchmark with the real-world scenarios and challenges.
To this end, we determine to track 5 kinds of moving objects of interest (\textit{i.e.}, person, animal, vehicle, aircraft, and boat) in 4 kinds of scenarios (\textit{i.e.}, hand-held, drone-mounted, vehicle-mounted, and video surveillance, as shown in Table~\ref{attribute-defination}).
Unlike RGB object tracking, which is interested in arbitrary objects, TIR object tracking is usually interested in objects with prominent thermal radiation as the 5 categories mentioned above.
The selected 5 categories of objects cover most of the targets of interest of TIR object tracking in the civilian field.

After determining the object classes and scenarios, we first search TIR videos on the Youtube website.
Unfortunately, unlike RGB videos, TIR videos are limited on Youtube.
We try our best to obtain 600 TIR videos and each of them is within ten minutes.
Since we use TIR videos with the white-hot style for tracking, we filter out some videos of other palette styles (\textit{e.g.}, iron, rainbow, and cyan).
We then convert the rest videos into image sequences and choose the fragments according to the following principles.
First, the object must be active, which is caused by the movement of itself or the camera.
Second, the time that the object is fully occluded or out of image range does not exceed one second.
Third, the length of the fragment does not exceed 4K frames.
In addition, we choose 150 TIR sequences from existing datasets, such as BU-TIV~\cite{TIV}, and RGB-T~\cite{RGB-T}.
Finally, we get 1,400 sequences that contain more than 600K frames and 47 object sub-classes of interest.

After getting all the sequences, we split them into a training dataset and an evaluation dataset.
We first choose 200 sequences, each of them contains at least one tracking challenging factor, for evaluation.
Then, we evaluate all the trackers using these sequences and then select the most difficult 120 sequences as the final evaluation subset according to the difficulty degree of each sequence, which is computed from the average success score of all the evaluated trackers.

\begin{figure}[t]
	\begin{center}
		\includegraphics[width=0.46\textwidth]{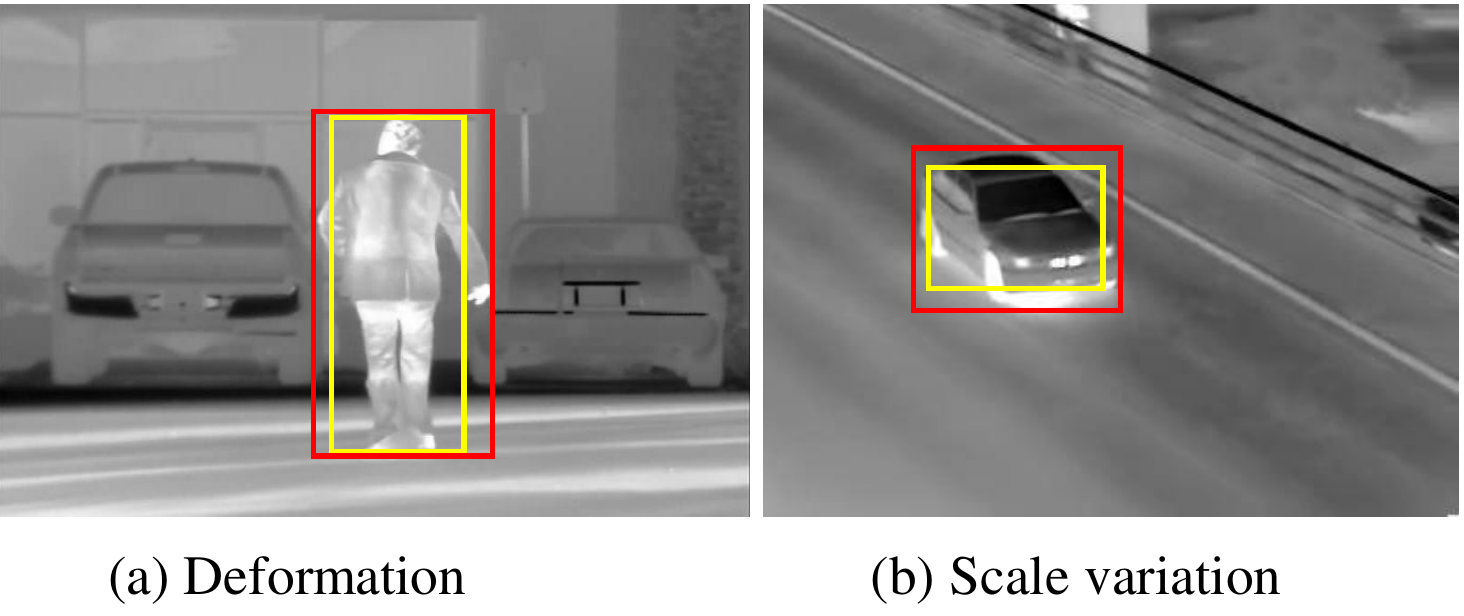}
	\end{center}
	\caption{Examples of the bounding box adjustment. The yellow bounding boxes are generated by using a tracking algorithm and the red bounding boxes are manually adjusted. }
	\label{calibration}
\end{figure}

\subsection{High-quality annotation}
\label{annoation}
We first decide an object of each frame contains 4 kinds of local annotation information, including object class, position, occlusion, and identity.
We use a straight minimum bounding box to record the position of an object.
When an object is occluded above 50\% or out of the image above 50\%, we define this occlusion as true.
This attribute can be used to exclude the obvious noise and is useful for training deep models.

Considering labeling is a time-consuming and labor-intensive task, we design an auxiliary label tool (see the \textbf{supplementary material}) based on the ECO-HC~\cite{ECO} tracker.
This label tool helps us generate a bounding box of the object in every frame semi-automatically.
When the tracker is set to track the object in a short time (\textit{e.g.}, within 10 frames), the generated bounding boxes are accurate in most situations.
However, when the object undergoes drastic appearance variation or scale change in a short time, the generated bounding boxes of the label tool are not quite accurate.
For these bounding boxes, we adjust them manually, as shown in Fig.~\ref{calibration}.
We suggest that the label tool makes the annotation more accurate, smoother, and faster than the annotation in each frame manually.

To complete the annotation, we assemble an annotation team comprised of 8 Ph.D. students with careful training.
To ensure the quality of the annotations, we verify the annotations frame by frame twice.
Eventually, we get 1,400 carefully annotated sequences. Some annotated sequences of the evaluation subset are shown in Fig.~\ref{evaluation-dataset}, and some annotated sequences of the training subset are shown in the \textbf{supplementary material}.

\begin{figure}[t]
	\begin{center}
		\includegraphics[width=0.46\textwidth]{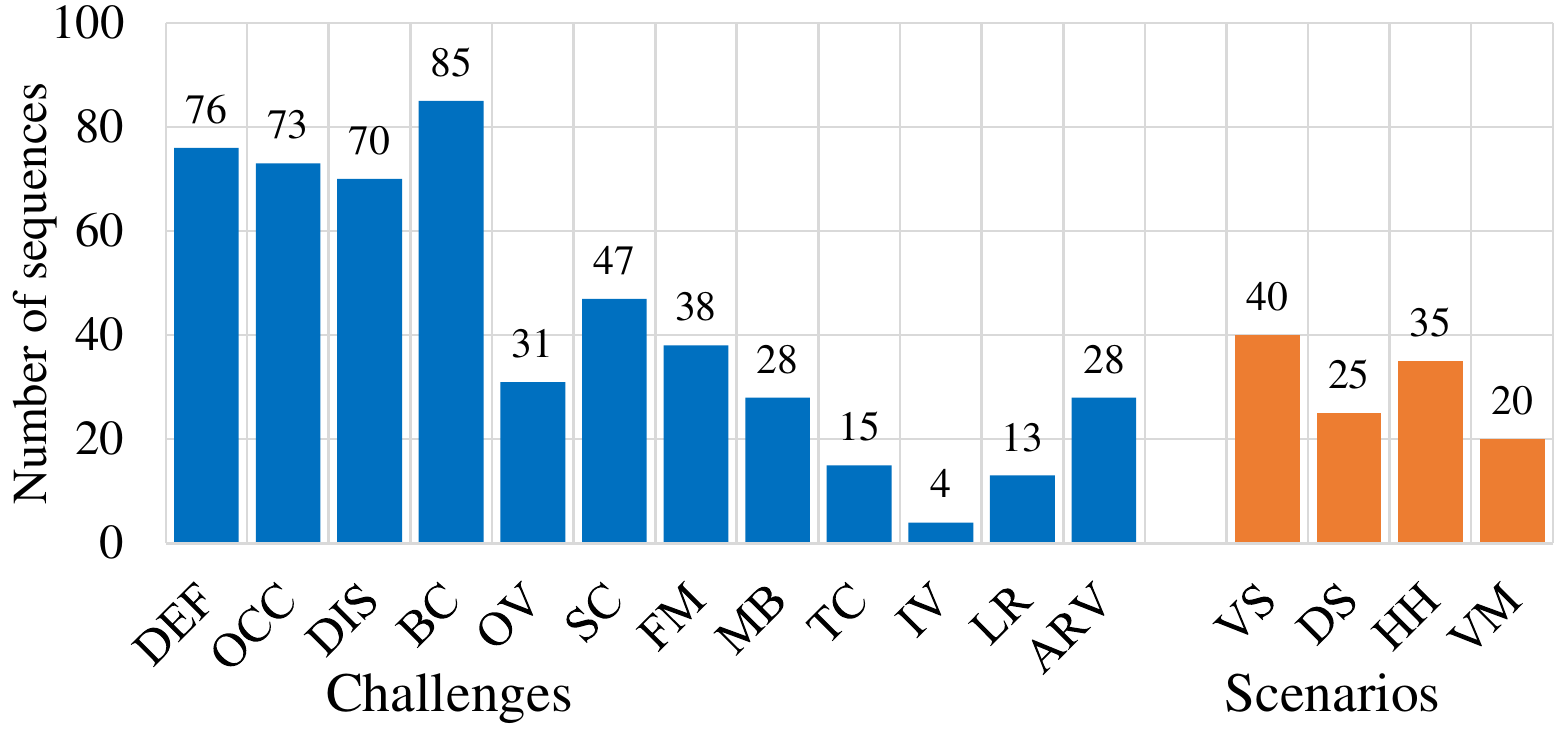}\vspace{-0.10in}
	\end{center}
	\caption{Attributes distribution of the evaluation subet of LSOTB-TIR.}
	\label{attribute-distribution}
\end{figure}

\begin{table*}[ht]
	\centering
	\caption{Comparison of tracking results of more than 30 trackers on LSOTB-TIR. We rank these trackers according to their success score. The property of a tracker includes feature representation (\textit{e.g.}, Deep: deep feature, HoG: histogram of gradient, Cova: covariance feature, CN: color name, Raw: raw pixel), search strategy (\textit{e.g.}, DS: dense search, RS: random search, PF: particle filter), category (\textit{e.g.}, D: discriminative, G: generative), and venue.  }
	\small
	\begin{tabular}{l|rrrr|rrrr}
		\toprule
		\multirow{2}{*}{Tracker} &  \multicolumn{4}{c|}{Performance} &  \multicolumn{4}{c}{Property} \\ \cline{2-9}
		& Success & Precision & Norm. Precision & Speed  & Representation & Search & Category & Venue \\
		\midrule
		ECO-TIR (Ours)     & \textcolor{red}{ \ding{172}} 0.631  & \textcolor{red}{\ding{172}} 0.768  &\textcolor{red}{\ding{172}} 0.695 & 18 fps  &Deep &DS  &D &-- \\
		ECO-stir~\cite{ECO-tir}  &\textcolor{green}{\ding{173}} 0.616 &\textcolor{green}{\ding{173}} 0.750 &0.672 &13 fps  &Deep &DS &D &TIP19 \\
		ECO~\cite{ECO}  &\textcolor{blue}{ \ding{174}} 0.609 &0.739 &0.670 &18 fps  &Deep &DS  &D &CVPR17 \\
		SiamRPN++~\cite{SiamRPN++} &0.604 &0.711 &0.651 &24 fps &Deep &DS &D &CVPR19 \\
		MDNet~\cite{MDNet}  &0.601 &\textcolor{green}{\ding{173}} 0.750 & \textcolor{green}{\ding{173}} 0.686 &1 fps  &Deep &RS &D &CVPR16 \\
		VITAL~\cite{VITAL}  &0.597 &\textcolor{blue}{ \ding{174}} 0.749 &\textcolor{blue}{ \ding{174}} 0.682 &3 fps  &Deep &RS &D &CVPR18 \\
		ATOM~\cite{ATOM} &0.595  &0.729 &0.647 &20 fps &Deep &RS &D &CVPR19 \\
		TADT~\cite{TADT}  &0.587 &0.710 &0.635 &40 fps  &Deep &DS &D &CVPR19 \\
		SiamMask~\cite{SiamMask} &0.579 & 0.705 &0.637 & 44 fps   &Deep &DS &D &CVPR19 \\
		ECO-HC~\cite{ECO}  &0.561 &0.690 &0.627 &27 fps  &HoG &DS &D &CVPR17 \\
		SiamFC-TIR (Ours)  & 0.554 & 0.700 &0.626 &45 fps &Deep &DS &D &-- \\
		BACF~\cite{BACF}  &0.535 &0.648 &0.591 &26 fps  &HoG &DS &D &ICCV17 \\
		SRDCF~\cite{SRDCF}  &0.529 &0.642 &0.574 &11 fps  &HoG &DS &D &ICCV15 \\
		UDT~\cite{UDT}  &0.523 &0.629 &0.575 &35 fps  &Deep &DS &D &CVPR19 \\
		MCCT~\cite{MCCT}  &0.522 &0.634 &0.574 &27 fps  &HoG\&CN &DS &D &CVPR18 \\
		SiamFC~\cite{Siamese-fc}  &0.517 &0.651 &0.587 &\textcolor{blue}{  \ding{174}} 45 fps  &Deep &DS &D &ECCVW16 \\
		SiamFC-tri~\cite{SiamFC-tri}  &0.513 &0.649 &0.583 &40 fps  &Deep &DS &D &ECCV18 \\
		CREST~\cite{CREST}  &0.504 &0.597 &0.544 &2 fps  &Deep &DS &D &ICCV17 \\
		Staple~\cite{Staple}  &0.492 &0.606 &0.548 &12 fps  &HoG\&CN &DS &D &CVPR16 \\
		MCFTS~\cite{MCFTS}  &0.479 &0.635 &0.546 &4 fps  &Deep &DS &D &KBS17 \\
		CFNet-TIR (Ours)    &0.478 & 0.580 &0.540 & 24 fps &Deep &DS &D &-- \\
		DSST~\cite{DSST}  &0.477 &0.555 &0.505 &\textcolor{green}{  \ding{173}} 50 fps  &HoG &DS &D &BMVC14 \\
		MLSSNet~\cite{MLSSNet}  &0.459 &0.596 &0.549 &25 fps  &Deep &DS &D &arXiv19 \\
		CFNet~\cite{CFNet}  &0.416 &0.519 &0.481 &24 fps  &Deep &DS &D &CVPR17 \\
		HSSNet~\cite{HSSNet}  &0.409 &0.515 &0.488 &15 fps  &Deep &DS &D &KBS19 \\
		HCF~\cite{HCF}  &0.404 &0.536 &0.485 &14 fps  &Deep &DS &D &ICCV15 \\
		HDT~\cite{HDT}  &0.403 &0.538 &0.478 &6 fps  &Deep &DS &D &CVPR16 \\
		TGPR~\cite{TGPR}  &0.403 &0.514 &0.495 &1 fps  &Cova &DS &D &ECCV14 \\
		RPT~\cite{RPT}  &0.388 &0.475 &0.427  &6 fps  &HoG &PF &D &CVPR15 \\
		Struck~\cite{struck}  &0.384 &0.477 &0.432  &17 fps  &Haar &DS &D &TPAMI15 \\
		DSiam~\cite{DSiam}  &0.380 &0.451 &0.393 &12 fps  &Deep &DS &D &ICCV17 \\
		L1APG~\cite{L1APG}  &0.371 &0.446 &0.424 &2 fps  &Raw &PF &G &CVPR12 \\
		LCT~\cite{LCT}  &0.364 &0.471 &0.430 &27 fps  &HoG\&Raw &DS &D &CVPR15 \\
		ASLA~\cite{ASLA}  &0.338 &0.429 &0.393 &3 fps  &Raw &PF &G &CVPR12 \\
		KCF~\cite{KCF}  &0.321 &0.418 &0.385  &\textcolor{red}{\ding{172}} 272 fps  &HoG &DS &D &TPAMI15 \\
		MIL~\cite{MIL}  &0.309 &0.378 &0.336 &19 fps  &Haar &DS &D &CVPR09 \\
		\bottomrule
	\end{tabular}\label{OPE-results-all}
\end{table*}

\subsection{Attribute definition}
\label{attributes}
In addition to the local attribute of each frame, we define two kinds of global attributes of a sequence in the evaluation dataset, namely scenario and challenge.
The corresponding attribute subsets can be used to further evaluate a tracker on specific attributes.
For the scenario attribute, we define 4 scenarios according to the TIR camera platform, including video surveillance, drone-mounted, hand-held, and vehicle-mounted, as shown in Table~\ref{attribute-defination}.
These scenario subsets can help us understand the strengths and weaknesses of a tracker on specific application scenarios.
For the challenge attribute, we define 12 challenges according to the real-world challenging factors in TIR videos.
For example, Thermal Crossover (TC) is defined as that two TIR objects with the same intensity cross each other and then lose their contour partly or fully.
Distractor (DIS) is defined as that the background near the target exists intra-class objects, which disturbs the tracker to recognize the tracking target.
This challenge is a frequent and serious problem in TIR object tracking.
Intensity Variation (IV) is defined as that the intensity of the target is changed due to its temperature variation or the brightness variation of the TIR camera.
This challenge is an unfrequent issue since the temperature of the target is stable in a short time.
Some other challenges, such as Background Clutter (BC), Deformation (DEF), Occlusion (OCC), Out of View (OV), Scale Change (SC), Fast Motion (FM),  Motion Blur (MB), Low Resolution (LR), and Aspect Ratio Variation (ARV), are defined in Table~\ref{attribute-defination}, and the distribution of each challenge is shown in Fig.~\ref{attribute-distribution}.

\section{Experiments}

\subsection{Evaluation criteria}
\label{criteria}
We use two widely used evaluation criteria in visual tracking, \textit{i.e.}, Center Location Error (CLE) and Overlap Ratio (OR), as the base metrics~\cite{OTB}. Base on these two metrics, precision, normalized precision, and success under One Pass Evaluation (OPE) are computed to measure the overall performance of a tracker.

\vspace{0mm}
\noindent{\textbf{Precision.}} CLE is the Euclidean distance between the center location of the predicted position and the ground-truth. Precision denotes the percentage of the successful frame whose CLE is within a given threshold (\textit{e.g.}, 20 pixels).

\vspace{0mm}
\noindent{\textbf{Normalized precision.}} Since the precision is sensitive to the resolution of the image and the size of the bounding box, we normalize the precision over the size of the ground-truth bounding box as that in TrackingNet~\cite{TrackingNet} and LaSOT~\cite{LaSOT}. We then use the Area Under Curve (AUC) of the normalized precision between 0 and 0.5 to rank the trackers.

\vspace{0mm}
\noindent{\textbf{Success.}} OR is the overlap rate between the predicted bounding box and the ground-truth. Success denotes the percentage of the successful frame whose OR is larger than a given threshold. We use a dynamic threshold [0 1], and the corresponding AUC is used to rank the tracking algorithms.

\vspace{0mm}
\noindent{\textbf{Speed.}} We use the average frame rate of a tracker on the dataset as the speed metric. We run all the trackers on the same PCs with an i7 4.0GHZ CPU and a GTX-1080 GPU.

\begin{figure*}[htbp]
	\centering
	\includegraphics[width=0.99\textwidth]{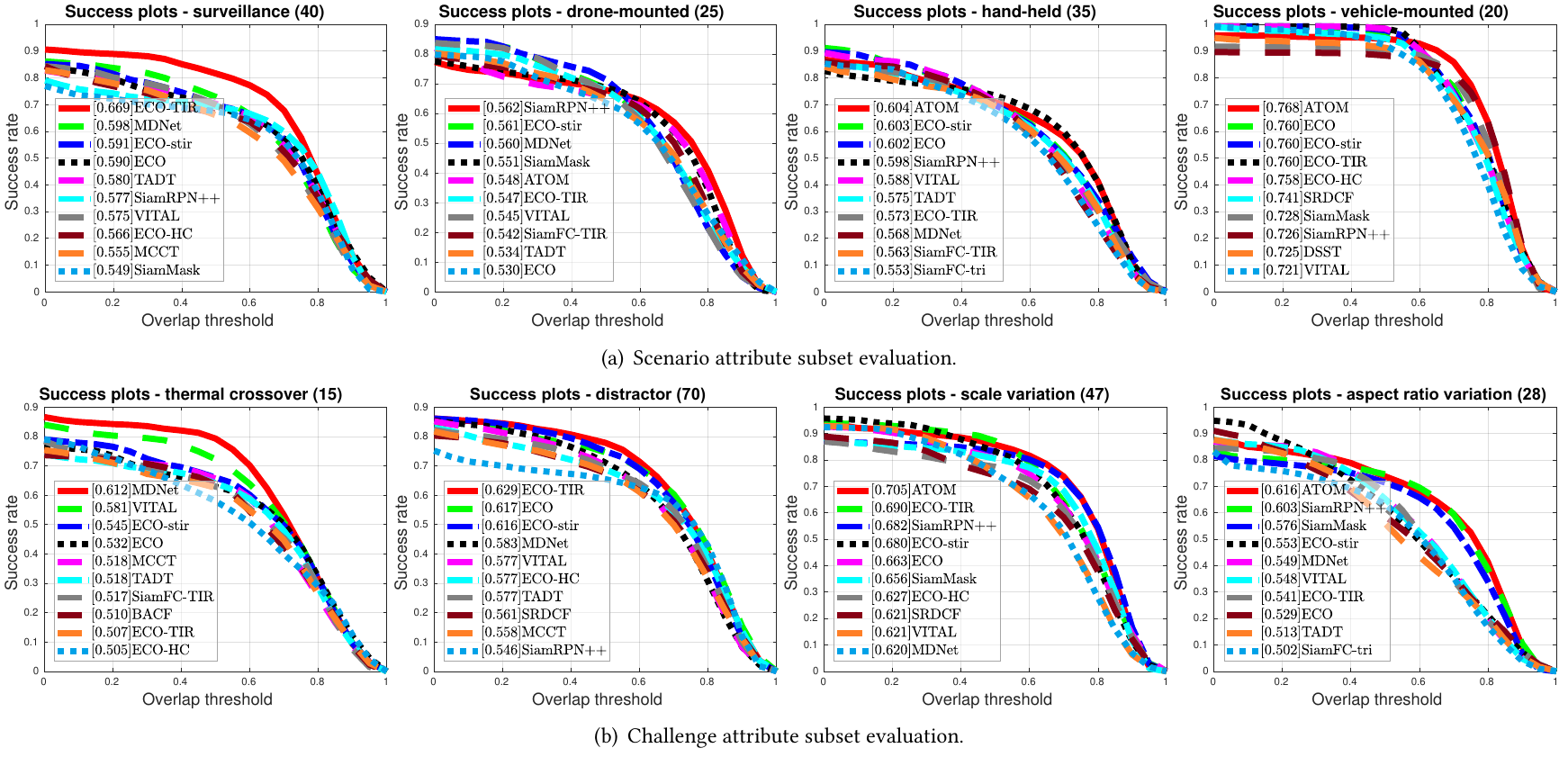}
	\caption{Attribute-based evaluation results. Only the top 10 trackers are shown for clarity. }
	\label{attribute-based-evaluation}
\end{figure*}

\subsection{Overall performance evaluation}
\label{overall-evaluation}
\noindent{\textbf{Evaluated trackers.}} We choose publicly available 33 TIR and RGB tracking methods for evaluation. These methods include sparse trackers, such as L1APG~\cite{L1APG}, and ASLA~\cite{ASLA}; correlation filter trackers, such as KCF~\cite{KCF}, DSST~\cite{DSST}, SRDCF~\cite{SRDCF}, BACF~\cite{BACF}, Staple~\cite{Staple}, ECO-HC~\cite{ECO}, and MCCT~\cite{MCCT}; other hand-crafted feature based trackers, including MIL~\cite{MIL}, TGPR~\cite{TGPR}, LCT~\cite{LCT}, Struck~\cite{struck}, and RPT~\cite{RPT}; deep feature based correlation filter trackers, such as HDT~\cite{HDT}, ECO~\cite{ECO}, HCF~\cite{HCF}, DeepSTRCF~\cite{DeepSTRCF}, MCFTS~\cite{MCFTS}, CREST~\cite{CREST}, and ECO-stir~\cite{ECO-tir}; matching based deep trackers, including SiamFC~\cite{Siamese-fc}, CFNet~\cite{CFNet}, DSiam~\cite{DSiam}, SiamFC-tri~\cite{SiamFC-tri}, TADT~\cite{TADT}, SiamRPN++~\cite{SiamRPN++}, SiamMask~\cite{SiamMask}, UDT~\cite{UDT}, HSSNet~\cite{HSSNet}, and MLSSNet~\cite{MLSSNet}; classification based deep trackers, such as MDNet~\cite{MDNet}, VITAL~\cite{VITAL}, and ATOM~\cite{ATOM}. We do not change the parameters of these trackers provided by the authors in the experiment. Furthermore, we re-train several deep trackers on the proposed TIR training dataset for evaluation, such as ECO-TIR, SiamFC-TIR, and CFNet-TIR. We use the backbone network of SiamFC-TIR as the feature extractor in the ECO-TIR tracker.

\vspace{0mm}
\noindent{\textbf{Results and analysis.}} Table~\ref{OPE-results-all} shows the overall performance and property of all the evaluated trackers.
Almost top 10 trackers are the deep feature based methods.
This shows that the deep feature is superior to the hand-crafted feature and deep trackers achieve promising performance in TIR object tracking.
ECO-TIR obtains the best success score (0.631) and precision score (0.768).
While ECO-stir~\cite{ECO-tir} using the synthetic TIR based deep feature obtains the second-best success score (0.616) and precision score (0.750).
Compared with ECO~\cite{ECO} which obtains the third-best success score (0.609) using the pre-trained RGB based deep feature, ECO-TIR and ECO-stir gain the success score by 2.2\% and 0.7\%, respectively.
This shows that the TIR based deep feature is superior to the RGB based deep feature in TIR object tracking.

Matching based deep trackers, such as SiamRPN++~\cite{SiamRPN++}, TADT~\cite{TADT}, SiamMask~\cite{SiamMask}, and SiamFC~\cite{Siamese-fc}, achieve comparable performance while running at a real-time speed.
These trackers are usually off-line trained to learn a matching network from a large-scale RGB dataset end-to-end.
Their favorable results demonstrate that the RGB based deep feature models of these trackers can represent TIR objects since there are some common patterns between RGB objects and TIR objects.
However, we argue that the RGB based deep feature has less discriminative capacity in representing TIR objects because the RGB based deep feature often tends to focus on the texture feature~\cite{geirhos2018imagenet}. Unlike RGB images, TIR images do not have color information and lack rich texture features. We suggest that the contour and structure features are critical for recognizing TIR objects. 
For example, compared with SiamFC~\cite{Siamese-fc}, our trained SiamFC-TIR achieves a 3.7\% success gain and a 3.9\% normalized precision gain.
This demonstrates that the learned TIR based deep feature has better discriminative capacity than the RGB based deep feature for distinguishing TIR objects.

Classification based deep trackers, including MDNet~\cite{MDNet}, VITAL~\cite{VITAL}, and ATOM~\cite{ATOM}, obtain favorable precision, which comes from the powerful discriminative capacity of the online learned binary classifier using the positive and negative samples of the tracked target.
This is important for a tracker to adapt the appearance variation of the tracked target.
Unfortunately, online training severely hampers their speed and easily leads to the over-fitting problem. However, for TIR trackers, we argue that online training is critical for more robust tracking. Because the online training can obtain a more powerful classifier to recognize subtle differences between the tracked object and distractors.

\subsection{Attribute-based evaluation}
\label{attribute-evaluation}
To understand the strengths and weaknesses of a tracker on special attributes, we evaluate all the trackers on the defined attribute subsets.
Fig.~\ref{attribute-based-evaluation}(a) shows the success scores of the top 10 trackers on the 4 scenario attribute subsets.
We notice that the ranking of these trackers is quite different between different scenario subsets.
For example, ECO-TIR is higher than MDNet~\cite{MDNet} by 7.1\% on the surveillance subset, while it is lower than MDNet by 1.3\% on the drone subset.
This shows that a tracker can not perform well on all the scenarios.
Furthermore, we find that the difficulty of the vehicle-mounted scenario subset is the smallest and all the top 10 trackers achieve good performance.
The major reason is that vehicle-mounted scenario subset contains fewer challenges and the limited tracked object classes.
These scenario subsets can help us develop specific-scenario based trackers to meet the requirements of real-world applications.
Fig.~\ref{attribute-based-evaluation}(b) shows the success scores of the top 10 trackers on 4 challenge attribute subsets.
MDNet gets the best success score (0.612) on the thermal crossover subset, which is higher than ECO-TIR by 10.5\%.
While ECO-TIR achieves the best success scores (0.629 and 0.621) on the distractor and background clutter subsets, which are higher than MDNet by 4.6\% and 2.7\% respectively.
This shows that a tracker can not handle all the challenges.
We attribute the good performance of ECO-TIR to the learned TIR based deep feature.
ATOM~\cite{ATOM} obtains the best success score on the scale variation and aspect ratio variation subsets.
This is because ATOM equips an overlap prediction network that can obtain a more accurate bounding box of the target.
More attribute-based results are shown in the \textbf{supplementary material}.

\begin{table}[t]
	\centering
	\caption{Comparison of 4 deep trackers trained on an RGB dataset and the proposed TIR training dataset, respectively. ``-TIR" denotes this tracker is trained on the proposed TIR training dataset.}
	\small
	\begin{tabular}{l|c|cc|cc}
		\toprule
		\multirow{2}{*}{Tracker} & VOT-TIR16~\cite{VOT-TIR2016} &\multicolumn{2}{c|}{PTB-TIR~\cite{PTB-TIR}}  &\multicolumn{2}{c}{LSOTB-TIR (Ours)} \\\cline{2-6}
		& EAO    & Pre.  & Suc.     & Pre.  & Suc.  \\
		\midrule
		SiamFC~\cite{Siamese-fc} &0.225 & 0.623 & 0.480    & 0.651 & 0.517 \\
		SiamFC-TIR &\textbf{0.250}  &\textbf{0.758}  &\textbf{0.566}    &\textbf{0.700}  & \textbf{0.554} \\
		\midrule
		CFNet~\cite{CFNet} &0.254  & 0.629 & 0.449    & 0.519 & 0.416 \\
		CFNet-TIR &\textbf{0.289}  & \textbf{0.726} & \textbf{0.530}    & \textbf{0.580} & \textbf{0.478} \\
		\midrule
		HSSNet~\cite{HSSNet} &0.262 & 0.689 & 0.468    & 0.515 & 0.409 \\
		HSSNet-TIR &\textbf{0.271}  & \textbf{0.723} & \textbf{0.490}    &\textbf{0.566}  &\textbf{0.435}  \\
		\midrule
		ECO~\cite{ECO} &0.267 &0.838 &0.633 &0.739 & 0.609 \\
		ECO-TIR &\textbf{0.290} &\textbf{0.858} &\textbf{0.650} &\textbf{0.768} & \textbf{0.631} \\
		\bottomrule
	\end{tabular}
	\label{validation-training-set}
\end{table}

\subsection{Training dataset validation }
\label{validation}
To validate that the proposed TIR training dataset can boost the performance of deep TIR trackers, we re-train 4 representative deep trackers using the proposed TIR training dataset.
Then, we compare them with the original trackers on 3 TIR object tracking benchmarks, including VOT-TIR16~\cite{VOT-TIR2016}, PTB-TIR~\cite{PTB-TIR}, and LSOTB-TIR.
Table~\ref{validation-training-set} shows that all the re-trained deep trackers achieve better performance.
Despite CFNet~\cite{CFNet} trained on a larger RGB dataset (VID~\cite{ILSVRC}), which is 4 times larger than the proposed TIR training dataset,  CFNet-TIR obtains an 8.1\% success gain on PTB-TIR and a 3.5\% Expected Average Overlap (EAO~\cite{VOTTIR2017}) gain on VOT-TIR16.
While SiamFC-TIR achieves a 13.5\% precision gain on PTB-TIR and a 3.7\% success gain on LSOTB-TIR.
ECO-TIR not only gets the best performance on LSOTB-TIR but also obtains, compared with ECO, a 2.3\% EAO gain on VOT-TIR16 and a 1.7\% success gain on PTB-TIR.
These results demonstrate that the proposed TIR training dataset significantly improves the performance of deep TIR trackers.

\section{Conclusions}
\label{conclusion}
In this paper, we propose a large-scale and high-diversity general thermal infrared object tracking benchmark with high-quality annotations, called LSOTB-TIR.
To the best of our knowledge, LSOTB-TIR is the largest and most diverse TIR object tracking benchmark to date.
By releasing LSOTB-TIR, we help the community develop deep learning based TIR trackers and evaluate them fairly and comprehensively.
We conduct extensive evaluation experiments with more than 30 trackers on LSOTB-TIR.
The results show that the deep trackers achieve promising performance. Furthermore, we re-train several deep trackers using the proposed training dataset, and their results demonstrate that the proposed training dataset significantly boosts the performance of deep TIR trackers. We suggest that learning TIR-specific deep feature for improving TIR object tracking is one of the main ways in the future. 

\section{Acknowledgment}
This study is supported by the National Natural Science Foundation of China (Grant No.61672183), by the Shenzhen Research Council (Grant No.JCYJ2017041310455226946, JCYJ20170815113552036).

\newpage
\bibliographystyle{ACM-Reference-Format}

\end{document}